\documentclass[conference]{IEEEtran}
\IEEEoverridecommandlockouts
\usepackage{cite}
\usepackage{amsmath,amssymb,amsfonts}
\usepackage{algorithmic}
\usepackage{graphicx}
\usepackage{textcomp}
\usepackage{xcolor}
\def\BibTeX{{\rm B\kern-.05em{\sc i\kern-.025em b}\kern-.08em
    T\kern-.1667em\lower.7ex\hbox{E}\kern-.125emX}}

\usepackage{algorithm}
\usepackage{algorithmic}
\usepackage{hyperref}
\usepackage{float}
\usepackage{multirow} 
\usepackage{comment}
\usepackage{booktabs}
\usepackage{subcaption}
\usepackage{amsthm}
\newtheorem{problem}{Problem}
\usepackage{ textcomp }
\usepackage[para]{footmisc}

\begin{document}

\newcommand{\todo}[1]{[\textcolor{red}{{TODO: }#1}]}
\newcommand{\note}[1]{[\textcolor{blue}{{NOTE: }#1}]}


\title{UGCE: User-Guided Incremental Counterfactual Exploration
\thanks{This paper has been accepted to the ForgtAI Workshop at IJCNN 2025.}
}

\author{\IEEEauthorblockN{Christos Fragkathoulas}
\IEEEauthorblockA{\textit{University of Ioannina and Archimedes,} \\
\textit{Athena Research Center, Greece}\\
ch.fragkathoulas@athenarc.gr}
\and
\IEEEauthorblockN{Evaggelia Pitoura}
\IEEEauthorblockA{\textit{University of Ioannina and Archimedes,} \\
\textit{Athena Research Center, Greece}\\
pitoura@uoi.gr}
}

\maketitle

\begin{abstract}

Counterfactual explanations (CFEs) are a popular approach for interpreting machine learning predictions by identifying minimal feature changes that alter model outputs. However, in real-world settings, users often refine feasibility constraints over time, requiring counterfactual generation to adapt dynamically. Existing methods fail to support such iterative updates, instead recomputing explanations from scratch with each change, an inefficient and rigid approach.
We propose \emph{User-Guided Incremental Counterfactual Exploration (UGCE)}, a genetic algorithm-based framework that incrementally updates counterfactuals in response to evolving user constraints.
Experimental results across five benchmark datasets demonstrate that UGCE significantly improves computational efficiency while maintaining high-quality solutions compared to a static, non-incremental approach. Our evaluation further shows that UGCE supports stable performance under varying constraint sequences, benefits from an efficient warm-start strategy, and reveals how different constraint types may affect search behavior.

\end{abstract}


\section{Introduction}
Machine learning (ML) models are increasingly deployed in high-stakes decision-making domains, including lending, college admissions, and hiring, where their predictions influence critical life outcomes \cite{siddiqi2012credit, waters2014grade, liem2018psychology}. However, these models often function as black boxes, making it difficult for stakeholders to understand the rationale behind predictions, particularly when an unfavorable decision is made. This lack of transparency has driven the need for explanation techniques that help users interpret and contest automated decisions \cite{lipton2018mythos, arrieta2020explainable}. As a result, research on explainability has gained significant traction, leading to a wide array of methodologies aimed at making ML models more transparent and interpretable \cite{molnar2020interpretable, 10.1145/3677119, bodria2023benchmarking, dwivedi2023explainable, arrieta2020explainable, adadi2018peeking, guidotti2024counterfactual, fragkathoulas2024explaining, prado2024survey}.

Among various explanation methods, counterfactual explanations (CFEs) have emerged as a prominent approach, as they provide suggestions by indicating minimal feature modifications that would have led to a different model outcome \cite{wachter2017counterfactual}. Counterfactuals not only explain a decision but also suggest actionable interventions for an individual to get a favorable outcome. For example, if a loan is denied, a counterfactual might state that the applicant would qualify with an MSc degree or a $10K$ income increase.

Despite their promise, the practical utility of CFEs is often limited by user dissatisfaction with the generated explanations. In real-world scenarios, users frequently begin with a minimal or incomplete set of constraints and only later realize that certain feature changes are unrealistic or undesirable. While several methods incorporate feasibility constraints \cite{poyiadzi2020face, mothilal2020explaining}, they assume a static specification of user preferences. As a result, they fall short when users seek to iteratively refine constraints—for instance, by disallowing changes to immutable features or tightening bounds on others.

To support such dynamic user needs, several recent efforts have introduced interactive \emph{user interfaces (UIs)} that allow users to refine constraints, adjust thresholds, or personalize counterfactual generation through visual and guided controls \cite{cheng2020dece, gomez2020vice, wang2023gam, wu2021polyjuice, wexler2019if, krause2016interacting}. While these systems enhance usability and transparency, they do not address the core optimization logic. Each new user update still triggers a full regeneration of counterfactuals from scratch, failing to reuse earlier search progress. As a result, these approaches, though interactive at the interface level, remain fundamentally \emph{static} in how they compute counterfactuals, limiting their scalability and responsiveness in iterative decision-making scenarios.

In reality, counterfactual exploration is inherently a dynamic process: users continuously refine constraints, shifting the feasible region with each interaction. This calls for optimization methods that can efficiently adapt to such evolving objectives. Genetic Algorithms (GAs) are particularly well-suited to this setting due to their flexibility in navigating constrained, non-convex spaces and their ability to maintain diverse candidate populations through stochastic operations. Moreover, extensive research in dynamic optimization has shown that GAs can respond effectively to environmental changes using mechanisms such as memory-based initialization, population repair, and targeted diversity injection, without restarting the search from scratch \cite{goldberg1987nonstationary, cobb1990investigation, grefenstette1992genetic, vavak1996comparative, branke1999memory, louis1996genetic, mori2001adaptation, ramsey1993case, yang2005memory}.

In this work, we introduce the \emph{User-Guided Incremental Counterfactual Exploration (UGCE)} framework, which enables true iterative counterfactual generation. UGCE dynamically updates newly imposed constraints and reuses the previously evolved population, avoiding full reinitialization. As the user refines feasibility constraints, UGCE selectively repairs individuals that violate the new requirements while preserving those that remain valid. This approach provides efficient, interaction-aware counterfactual refinement that complements existing user-facing tools and brings optimization in line with real-world workflows.


The remainder of the paper is structured as follows. Section~\ref{sec:problem_definition} formalizes the problem of counterfactual generation under dynamic user constraints. Section~\ref{sec:ugce} introduces the UGCE framework.
Section~\ref{sec:experiments} presents evaluation across benchmark datasets, analyzing performance under various user interaction scenarios. Section~\ref{sec:relatedwork} discusses related work, Section~\ref{sec:summary} concludes, and section \ref{sec:future_directions} presents directions for future research.

\section{Problem Definition}
\label{sec:problem_definition}

We consider a binary classifier \( f: \mathbb{R}^d \to \{0, 1\} \), which maps instances from a \(d\)-dimensional feature space into two classes. 
Let \(\mathbf{x} \in \mathbb{R}^d\) denote an original instance predicted as \(f(\mathbf{x}) = y\).
Given a prediction model \(f\) and an input instance \(\mathbf{x} \in \mathbb{R}^d\), a \emph{counterfactual explanation} is an alternative instance \(\mathbf{x}' \in \mathbb{R}^d\) such that \(f(\mathbf{x}') \neq f(\mathbf{x})\), and \(\mathbf{x}'\) remains similar to \(\mathbf{x}\) under some meaningful cost measure.
The goal is to identify a counterfactual \(\mathbf{x}'\) that minimally deviates from \(\mathbf{x}\) while changing the model prediction.
This is framed as an optimization problem \cite{wachter2017counterfactual}:
\[
\begin{aligned}
\mathbf{x}' &= \arg\min_{\mathbf{x}'} \ \text{cost}(\mathbf{x}, \mathbf{x}') 
& \text{s.t.} \quad &f(\mathbf{x}') \neq f(\mathbf{x}),
\end{aligned}
\]
where \(\text{cost}\) can be, for example, the \(L_2\) norm or a user-defined dissimilarity measure.
Beyond minimality, real-world counterfactuals are often required to be \emph{feasible}. For instance, they should not propose changes to immutable attributes like gender or race \cite{wachter2017counterfactual}. 
To formalize feasibility, we define a constraint set \(\mathcal{C}\), where each constraint \(c_i \in \mathcal{C}\) applies to a specific feature \(i \in \{1, \ldots, d\}\). 
To avoid conflicting specifications, we assume that at most one constraint is active per feature at any given time. 
Given a set \(\mathcal{C}\), we say that a counterfactual \(\mathbf{x}'\) is feasible for \(\mathbf{x}\) under \(\mathcal{C}\) if it satisfies all constraints in \(\mathcal{C}\). In this work, we consider three types of constraints:
\begin{itemize}
    \item \textbf{Immutability constraint:} Feature \(x_i\) must remain unchanged, i.e., \(x_i = x'_i\).
    \item \textbf{Range constraint:} Feature \(x'_i\) must lie within a valid interval \([a_i, b_i]\), where $a_i, b_i$ are specified externally (e.g., by the user or the data domain).
    \item \textbf{Directionality constraint:} Feature \(x'_i\) must only increase or only decrease relative to \(x_i\).
\end{itemize}

The counterfactual generation under feasibility constraints becomes:
\[
\mathbf{x}' = \arg\min_{\mathbf{x}' \in \mathcal{F}(\mathcal{C})} \ \text{cost}(\mathbf{x}, \mathbf{x}')
\quad \text{s.t.} \quad f(\mathbf{x}') \neq f(\mathbf{x}),
\]
where \(\mathcal{F}(\mathcal{C})\) denotes the set of feasible counterfactuals that satisfy all constraints in \(\mathcal{C}\).

\noindent In interactive settings, a user may alter a given set \(\mathcal{C}\) of feasibility constraints to create a new set \(\mathcal{C}'\) by:
(a) deleting constraints in \(\mathcal{C}\), (b) adding new constraints, or (c) modifying existing ones.

\begin{problem}[]
Given \(\mathbf{x}\), a counterfactual \(\mathbf{x}'\) feasible under a constraint set \(\mathcal{C}\) for $\mathbf{x}$, and a new user-issued update resulting in an updated constraint set \(\mathcal{C}'\), find a new counterfactual \(\mathbf{x}''\) such that:
\[
f(\mathbf{x}'') \neq f(\mathbf{x}), \quad \mathbf{x}'' \in \mathcal{F}(\mathcal{C}').
\]
\end{problem}
We envision a system where an initial counterfactual is presented to the user, who then iteratively refines the explanation by modifying the constraint set. Each update alters the feasible region, prompting the algorithm to adjust the counterfactual accordingly.

Formally, let \(\mathcal{C}_t\) be the set of constraints imposed at iteration \(t\), and let \(\mathbf{x}'^{(t)}\) be a counterfactual satisfying:
\[
f(\mathbf{x}'^{(t)}) \neq f(\mathbf{x}), \quad \mathbf{x}'^{(t)} \in \mathcal{F}(\mathcal{C}_t).
\]

At iteration \(t+1\), the user provides an update that results in a modified constraint set \(\mathcal{C}_{t+1}\). The system must then compute a new counterfactual \(\mathbf{x}'^{(t+1)}\) such that:
\[
f(\mathbf{x}'^{(t+1)}) \neq f(\mathbf{x}), \quad \mathbf{x}'^{(t+1)} \in \mathcal{F}(\mathcal{C}_{t+1}).
\]

\section{UGCE}
\label{sec:ugce}
\begin{figure*}[!ht]
    \centering
    \includegraphics[width=1\linewidth]{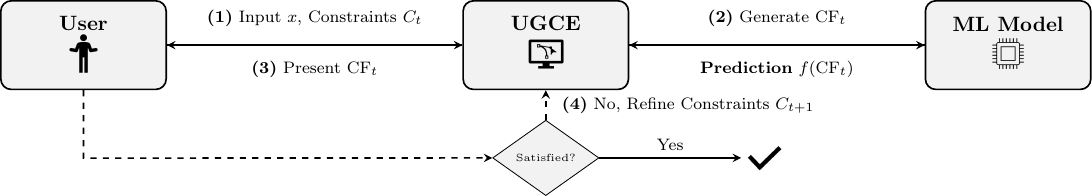}
    \caption{UGCE Pipeline}
    \label{fig:ugce_pipeline}
\end{figure*}

In this section, we present the \textbf{User-Guided Counterfactual Exploration (UGCE)} framework—our solution to the problem defined in Section~\ref{sec:problem_definition}. UGCE is designed to address the dynamic user interaction in counterfactual explanation, where users iteratively refine their feasibility constraints across multiple steps. This dynamic setting is illustrated in the abstract pipeline of Fig.~\ref{fig:ugce_pipeline}, where a user sets constraints, receives a counterfactual from the system, and may revise their constraints in response to the generated explanation.

While several methods have been proposed for counterfactual generation \cite{guidotti2019factual,mothilal2020explaining,sharma2020certifai,kanamori2020dace,ustun2019actionable,goethals2024precof,russell2019efficient,poyiadzi2020face,guyomard2023generating}, they assume a static setting where all constraints are imposed upfront and remain fixed. These methods solve the problem in a single step, ignoring the iterative nature of user involvement.

To support dynamic, user-in-the-loop interaction, UGCE employs \textit{Genetic Algorithms (GAs)} as its optimization engine. GAs have previously been used for counterfactual generation under fixed constraints \cite{mothilal2020explaining, sharma2020certifai} and are particularly well-suited for this problem due to their ability to explore non-convex, constrained spaces and efficiently evolve candidate solutions using stochastic operators like selection, crossover, and mutation.

However, dynamic systems require more than static feasibility handling, they must also be \textit{responsive}. That is, they should react promptly to user feedback without retraining or recomputing from scratch. To achieve this, UGCE introduces an incremental update mechanism that builds upon previous search states rather than restarting.
This is critical in interactive settings where low latency and continuity across iterations are necessary for practical deployment.

Prior work in dynamic optimization has shown that GAs can adapt effectively to changing environments by leveraging strategies such as memory reuse and population repair \cite{branke2003designing}. UGCE extends these ideas to the user-driven counterfactual setting by reusing the population from previous iterations and repairing candidate solutions that violate newly imposed constraints. This warm-start strategy enables efficient, real-time updates in evolving feasible spaces.

\subsection{Framework Overview}
UGCE operates through a sequence of user-in-the-loop optimization rounds. At each round \(t\), a counterfactual \( \mathbf{x}'^{(t)} \) is generated given the current constraint set \(\mathcal{C}_t\), and presented to the user. If the user accepts the counterfactual, the process terminates. Otherwise, the user modifies the constraint set (e.g., adds a directionality constraint), forming \(\mathcal{C}_{t+1}\). The algorithm then continues the counterfactual search using the evolved population from iteration \(t\), now filtered to conform to \(\mathcal{C}_{t+1}\).

This workflow leads to the following iteration loop:
\begin{itemize}
    \item Generate candidate counterfactuals.
    \item Present the best candidate to the user.
    \item If rejected, update the constraint set based on user feedback.
    \item Reuse and repair the population to reflect the new constraints.
    \item Resume the search from the adapted population.
\end{itemize}

\subsection{Evolutionary Methodology}
UGCE applies standard genetic algorithm (GA) operators. 
GAs are optimization techniques inspired by natural evolution, first introduced by Holland \cite{holland1975adaptation}. They work by maintaining a population of candidate solutions based on the original instance \(\mathbf{x}\) and iteratively improving them over generations. 
In each iteration, better-performing candidates are kept, while new ones are generated by combining or modifying them, following the basic idea that better solutions are more likely to survive and evolve.
In UGCE, this evolutionary process is guided by the user: new constraints may be introduced based on user feedback, and the population is updated accordingly. The evolutionary steps of the UGCE are detailed below:

\emph{Step 1: Initialization.} 
At the first iteration, UGCE supports two strategies for initializing the population:
(i) retrieving \(k\)-nearest neighbors from the opposite class, drawn from the training data (using a KD-Tree structure \cite{Friedman1977AnAF}), or
(ii) generating a diverse population of synthetic counterfactuals around \(\mathbf{x}\) that respect the initial constraint set \(\mathcal{C}_0\).
In subsequent iterations, the population is inherited from the previous round and repaired, if necessary, to satisfy the updated constraint set \(\mathcal{C}_t\).

\emph{Step 2: Fitness Evaluation.}
Each candidate from the population is scored using a three-objective fitness function.
Specifically, it jointly optimizes: altering the model prediction, maintaining proximity to the original instance, and minimizing the number of changed features, defined as:
\begin{gather}
\text{Fitness}(\mathbf{x}, \mathbf{x}') = 
- \lambda_1 \cdot \underbrace{\frac{\sum_{i \in d} w_i |x_i' - x_i|}{\sum_{i \in d} w_i}}_{\text{$Proximity$}} \notag \\
- \lambda_2 \cdot \underbrace{\sum_{i=1}^{d} \mathbb{I}(|x_i - x'_i| > \epsilon)}_{\text{$Sparsity$}} 
+ \lambda_3 \cdot \underbrace{L_{\text{pred}}(\mathbf{x}')}_{\text{$Prediction$}}    
\end{gather}
where, 
\(\lambda_1, \lambda_2, \lambda_3 \in \mathbb{R}^+\) are hyperparameters that control the relative importance of each objective, and \(w_i\) are feature-level weights (e.g., inverse MAD or standard deviation). \(\mathbb{I}(\cdot)\) is an indicator function that computes the number of significantly changed features under a user-defined threshold \(\epsilon > 0\), and the prediction score is defined as:
\begin{align}
L_{\text{pred}}(\mathbf{x}') = 
\begin{cases} 
\alpha, & \text{if } f(\mathbf{x}') \neq f(\mathbf{x}) \\ 
-\beta, & \text{otherwise}, \notag
\end{cases}    
\end{align}
where \(\alpha > 0\) and \(\beta > 0\) are reward and penalty values, respectively, controlling the pressure towards flipping the decision of the classifier.

\emph{Step 3: Selection.}
Candidates are sampled for reproduction based on their fitness scores. By default, we use Stochastic Universal Sampling (SUS), which balances exploration and exploitation by offering fairer chances to mid-range candidates compared to more greedy methods like tournament selection \cite{baker1987reducing}. 
This operation is crucial to avoid \textit{premature convergence} or having a population dominated by one type of solution \cite{greenstein2023best, douguet2000genetic, pereira2005effect}.

\emph{Step 4: Crossover and Mutation.}
Selected candidates are recombined and mutated to explore the neighborhood around promising solutions. These genetic operators introduce variability while maintaining feasibility under the current constraint set.

\emph{Step 5: User Feedback and Adaptation.}
Once the search converges or reaches a stopping criterion, the best candidate is presented to the user. If accepted, the process terminates. Otherwise, the user updates the constraint set, and UGCE resumes the search from the current population, repaired as needed, under the new constraints.

The above steps are shown in Algorithm \ref{alg:algorithm_1}.

\begin{algorithm}[ht]
\caption{UGCE: User-Guided Counterfactual Exploration}
\begin{algorithmic}[1]
\STATE Initialize population \(\mathcal{P}_0\) from \(\mathbf{x}\) under constraints \(\mathcal{C}_0\)
\FOR{iteration \(t = 0, 1, \dots\)}
    \REPEAT
        \STATE Evaluate fitness of each $\mathbf{x}' \in \mathcal{P}_t$
        \STATE Select parents (e.g., SUS, tournament)
        \STATE Apply crossover and mutation
    \UNTIL{convergence or early stopping}
    \STATE Present best candidate $\mathbf{x}'^{(t)}$ to user
    \IF{user accepts}
        \STATE \textbf{return} $\mathbf{x}'^{(t)}$
    \ELSE
        \STATE User updates constraints $\mathcal{C}_{t+1}$
        \STATE Repair $\mathcal{P}_t$ to satisfy $\mathcal{C}_{t+1}$ and set $\mathcal{P}_{t+1} \leftarrow \mathcal{P}_t$
    \ENDIF
\ENDFOR
\end{algorithmic}
\label{alg:algorithm_1}
\end{algorithm}

\section{Experimental Evaluation}
\label{sec:experiments}
We conduct four sets of experiments to evaluate the effectiveness of UGCE. Our experiments focus on (i) A comparison between UGCE and a static state-of-the-art approach (DiCE) that uses Genetic Algorithms to generate counterfactuals under fixed constraints, (ii) Investigating the benefit of the \textit{fixing-violators} incremental warm-start strategy over a \textit{random warm-start baseline} in the incremental settings, and (iii) studying the stability of UGCE under different sequences to evaluate its behavior across diverse user interaction patterns, and (iv) A preliminary analysis of the influence of each individual constraint type (immutability, range, directionality) on the counterfactual search process.

For the evaluations, we conduct experiments on five benchmark datasets.
The \texttt{German Credit}\footnote{\href{http://archive.ics.uci.edu/ml/machine-learning-databases/statlog/german/german.data}{German Credit Dataset}} contains financial and demographic attributes used to classify individuals as good or bad credit risks.
The \texttt{HELOC}\footnote{\href{https://www.kaggle.com/datasets/averkiyoliabev/home-equity-line-of-creditheloc}{HELOC Dataset}} includes credit account information to assess the risk of default on home equity lines of credit.
The \texttt{COMPAS}\footnote{\href{https://raw.githubusercontent.com/propublica/compas-analysis/bafff5da3f2e45eca6c2d5055faad269defd135a/compas-scores-two-years.csv}{COMPAS Dataset}} contains records used to predict recidivism risk in the criminal justice system.
The \texttt{Adult}\footnote{\href{https://archive.ics.uci.edu/dataset/2/adult}{Adult Dataset}} includes census data used to predict whether an individual earns more than \$50,000 annually.
\texttt{AdultCA}\footnote{\href{https://github.com/socialfoundations/folktables}{AdultCA Dataset}}, where we select data from 2023, as it represents the most up-to-date information available. This also consists of records of individuals used for predicting if their annual income exceeds $\$50,000$. 

For all experiments, we train a Random Forest classifier on each dataset using an 80\% training and 20\% test split. The classifier is used as a black-box model whose predictions are explained through counterfactuals. We focus on explaining the instances predicted with the negative (unfavorable prediction) class. Table~\ref{tab:dataset_stats} summarizes the classification accuracy on the test set, the number of categorical and numerical features, and the number of negative predictions per dataset.
To enable fair comparison with prior work, we adopt the same default hyperparameters as DiCE \cite{mothilal2020explaining}. 
In particular, we use: (a) fitness weights \(\lambda_1 = 0.2\), \(\lambda_2 = 0.2\), \(\lambda_3 = 1.0\), and (b) a binary reward in the prediction score with \(\alpha = 1\), \(\beta = 1\), resulting in \(+1\) for successful counterfactual flips and \(-1\) otherwise.
We select the KD-Tree strategy for population initialization in the first iteration.
All experiments were run on a machine equipped with an AMD Ryzen 9 5950X 16-core processor. The source code is available online\footnote{\href{https://github.com/xristosfrag/UGCE-User-Guided-Counterfactual-Exploration}{Project Repository}}.
\begin{table}[ht]
\centering
\caption{Dataset statistics and classifier performance.}
\label{tab:dataset_stats}
\begin{tabular}{lcccc}
\toprule
\textbf{Dataset} & \textbf{Accuracy} & \textbf{Categorical} & \textbf{Numerical} & \textbf{Negatives} \\
\midrule
\shortstack{\texttt{German}\\\texttt{Credit}} & 0.75 & 13 & 7  & 166   \\
\texttt{HELOC}         & 0.69 & 0  & 23 & 368   \\
\texttt{COMPAS}        & 0.63 & 3  & 5  & 550    \\
\texttt{Adult}         & 0.84 & 0  & 12 & 2,063 \\
\texttt{AdultCA}       & 0.80 & 0  & 9  & 20,203 \\
\bottomrule
\end{tabular}
\end{table}

\subsection{Counterfactual Update Methods}
We distinguish between two methods for generating counterfactuals under evolving user constraints:
\begin{itemize}
    \item \textbf{Baseline:} A counterfactual is computed in a single pass using all current constraints. If the user later modifies any constraint, the generation process restarts entirely from the original instance, discarding any prior solutions.

    \item \textbf{Incremental:} Counterfactuals are constructed iteratively. Starting from an existing solution, new constraints are integrated by adjusting the current counterfactual rather than restarting, preserving feasible modifications.
\end{itemize}

\subsection{Baseline Comparison}
In this experiment, we compare \textbf{UGCE-Baseline}, which recomputes counterfactuals from scratch at each step, and \textbf{UGCE-Incremental}, which computes counterfactuals iteratively.
We also use \textbf{DiCE-Baseline} 
that use DiCE \cite{mothilal2020explaining} for computing counterfactuals. 

The comparison evaluates several key metrics: computational time (\textbf{Time}) for all instances under explanation, the \textbf{Gens} (average number of generations until convergence), the average percentage of instances for which a valid counterfactual was found (\textbf{CF (\%)}), and quality-related measures such as \textbf{$Proximity$} (average proximity of counterfactual to the original instance), and average normalized \textbf{$Sparsity$} (average normalized number of feature changes). 


All reported values in Table~\ref{tab:baseline-comparison} are computed by averaging across all negatively predicted instances, each executed $5$ times to account for the stochastic nature of the evolutionary process. Standard deviations are included to assess the performance deviations of each method. We omit DiCE standard deviation results for the \texttt{AdultCA} dataset, as its runtime exceeded $22$ hours per run, making repeated executions computationally prohibitive
We use a fixed set of generic feasibility constraints across all datasets, disallowing unrealistic or non-actionable changes such as modifications to immutable attributes like race and sex. We focus solely on immutability constraints in this experiment to ensure fair comparison since DiCE does not natively support directionality constraints.

Several insights can be drawn from the results. First, both UGCE methods show significantly lower variability in their results compared to DiCE, despite both relying on evolutionary search. This increased stability makes UGCE more reliable in practical settings.
For instance, UGCE exhibits near-zero variability in computational time across runs, remaining consistently low, while DiCE shows substantial fluctuations at approximately $10$ seconds for \texttt{Adult} and $15$ seconds for \texttt{German Credit}.
More notably, the \textbf{CF (\%)} success rate of DiCE on \texttt{German Credit} and \texttt{Adult} has a standard deviation close to $40$ and $27$, respectively, indicating a lack of consistency and reliability in its ability to generate counterfactuals across repeated runs.

In terms of runtime, UGCE-Baseline generally outperforms DiCE-Baseline, with minor exceptions on the HELOC dataset (slightly slower) and German Credit dataset (where DiCE appears faster but achieves a significantly lower success rate). 
Notably, DiCE requires considerably more time on fairness-related datasets such as \texttt{COMPAS}, \texttt{Adult}, and \texttt{AdultCA}. DiCE-Baseline did not complete execution on the COMPAS dataset within a reasonable timeframe (exceeding 24 hours), hence its results are omitted. 
This difference is likely due to the continuous hinge loss function of DiCE \cite{mothilal2020explaining}, which can lead to slower convergence compared to the faster filtering via a discrete prediction reward of UGCE. As expected, UGCE-Incremental is consistently the fastest method, offering substantial runtime reductions essential for real-time, user-guided exploration.
%

Looking at the counterfactual success rate (\textbf{CF (\%)}), UGCE-Baseline consistently surpasses DiCE. While UGCE-Incremental has slightly lower success rates, due to the difficulty of escaping local optima after constraint updates, it still achieves high coverage in $3$ out of $5$ datasets, maintaining success rates above $79\%$ with a significant speedup.

Regarding \textbf{$proximity$} and \textbf{$sparsity$} metrics, which directly impact user interpretability, UGCE-Baseline yields the most minimal and least sparse counterfactuals. UGCE-Incremental follows closely, showing comparable quality despite the trade-off in optimization time. Overall, all three methods produce counterfactuals with low \textbf{$proximity$} and \textbf{$sparsity$} values, indicating generally high-quality explanations.

These results collectively demonstrate that UGCE provides more stable and efficient counterfactuals while achieving comparable solution quality in dynamic user-driven settings, with significant speedups.
\begin{table*}[ht]
\centering
\caption{
UGCE Methods and DiCE Comparison.}
\label{tab:experiment_results}
\begin{tabular}{l c | l | c c c c c c}
\toprule
\textbf{Dataset} & \textbf{Instances} & \textbf{Method}
& \textbf{Time} & \textbf{Gens}
& \textbf{CFs (\%)} & \textbf{$Proximity$} & \textbf{$Sparsity$} \\
\midrule

\multirow{3}{*}{\shortstack{\texttt{German}\\\texttt{Credit}}} & \multirow{3}{*}{$166$}
& UGCE-Baseline  & $22.44$ (s) (\textpm $0.15$)  & $6.22$ (\textpm $0.07$) & $98.81$ (\textpm $0.00$) & $0.08$ (\textpm $0.00$) & $0.02$ (\textpm $0$) \\
& & UGCE-Incremental  & $18.27$ (s) (\textpm $0.68$)  & $6.03$ (\textpm $0.03$) & $95.36$ (\textpm $0.69$) & $0.11$ (\textpm $0.01$) & $0.02$ (\textpm $0$) \\
& & DiCE-Baseline   & $18.61$ (s) (\textpm $15.41$) & $7.54$ (\textpm $1.29$) & $52.41$ (\textpm $39.63$) & $0.08$ (\textpm $0.01$) & $0.02$ (\textpm $0$) \\

\midrule
\multirow{3}{*}{\texttt{HELOC}} & \multirow{3}{*}{$365$}
& UGCE-Baseline  & $36.55$ (s) (\textpm $0.22$) & $6.00$ (\textpm $0.0$) & $99.46$ (\textpm $0$) & $0.03$ (\textpm $0$) & $0.03$ (\textpm $0$) \\
& & UGCE-Incremental   & $31.24$ (s) (\textpm $0.34$) & $6.00$ (\textpm $0.0$) & $92.43$ (\textpm $0$) & $0.03$ (\textpm $0$) & $0.03$ (\textpm $0$) \\
& & DiCE-Baseline   & $32.12$ (s) (\textpm $2.72$) & $6.91$ (\textpm $1.1$) & $100.0$ (\textpm $0$) & $0.06$ (\textpm $0$) & $0.03$ (\textpm $0$) \\

\midrule
\multirow{3}{*}{\texttt{COMPAS}} & \multirow{3}{*}{$505$}
& UGCE-Baseline  & $39.42$ (s) (\textpm $1.79$) & $6.02$ (\textpm $0$) & $93.11$ (\textpm $0$) & $0.02$ (\textpm $0$) & $0.02$ (\textpm $0$) \\
& & UGCE-Incremental   & $28.82$ (s) (\textpm $1.42$) & $8.10$ (\textpm $0.33$) & $47.86$ (\textpm $0.68$) & $0.05$ (\textpm $0$)& $0.03$ (\textpm $0$) \\
& & DiCE-Baseline   & $-$  & $-$    & $-$     & $-$ & $-$ \\

\midrule
\multirow{3}{*}{\texttt{Adult}} & \multirow{3}{*}{$2,049$}
& UGCE-Baseline  & $4.17$ (m) (\textpm $0.18$) & $6.25$ (\textpm $0$) & $99.56$ (\textpm $0$) & $0.03$ (\textpm $0$) & $0.02$ (\textpm $0$) \\
& & UGCE-Incremental   & $2.67$ (m) (\textpm $0.07$) & $6.16$ (\textpm $0.06$) & $79.55$ (\textpm $0.17$) & $0.04$ (\textpm $0$) & $0.02$ (\textpm $0$) \\
& & DiCE-Baseline   & $32.95$ (m) (\textpm $10.22$) & $18.75$ (\textpm $1.40$) & $86.18$ (\textpm $27.62$) & $0.04$ (\textpm $0$) & $0.02$ (\textpm $0$) \\

\midrule
\multirow{3}{*}{\texttt{AdultCA}} & \multirow{3}{*}{$20,203$}
& UGCE-Baseline  & $58.35$ (m) (\textpm $2.12$) & $6.04$ (\textpm $0.0$) & $99.99$ (\textpm $0.0$) & $0.05$ (\textpm $0.0$) & $0.05$ (\textpm $0.0$) \\
& & UGCE-Incremental   & $21.62$ (m) (\textpm $0.89$) & $6.02$ (\textpm $0.0$) & $57.35$ (\textpm $0.3$) & $0.07$ (\textpm $0.0$)& $0.06$ (\textpm $0.0$) \\
& & DiCE-Baseline   & $22.29$ (h)   & $6.04$  & $1.12$ & $0.07$ & $0.04$ \\
\bottomrule
\end{tabular}
\label{tab:baseline-comparison}
\end{table*}

\subsection{Warm Start Strategies Analysis}
To assess our design choice of repairing the existing population when new constraints are imposed, we compare the default \emph{violator-fixing} mechanism of UGCE to a randomized alternative. Specifically, instead of modifying only the individuals that violate the new constraints, the alternative reinitializes the population by sampling randomly around the instance under explanation while still respecting the updated constraints. Both strategies serve as warm starts for the new search, differing only in how they respond to user-driven constraint changes.

Table~\ref{tab:fix_vs_random} reports the average results and total time required over five runs per dataset. Following earlier experiments, we omit standard deviations as UGCE exhibits negligible variance. We also report p-values from two-sided t-tests to assess whether the performance differences between the two strategies are statistically significant.

Across all five datasets, the \textbf{violator-fixing} strategy consistently outperforms the \textbf{random restart} alternative in both \emph{computational efficiency} and \emph{counterfactual success rate} (CFEs \%). 
On quality metrics (\textbf{$Proximity$}, \textbf{$Sparsity$}), the fixing strategy also yields significantly better results in all datasets except \texttt{AdultCA}, where both methods perform comparably. This reinforces the advantage of retaining partially valid individuals rather than reinitializing the entire population, a key strength of the incremental design of UGCE.

Overall, these findings empirically validate our choice to repair rather than restart the population. Given its consistent performance gains and statistically significant advantages (as shown by the reported p-values), we adopt the violator-fixing strategy as the default warm-start method in UGCE.
\begin{table}[ht]
    \centering
    \caption{Fix and Random Warm Start Strategies Comparison for UGCE-Incremental.}
    \scriptsize
    \setlength{\tabcolsep}{2pt}
    \begin{tabular}{lcccccc}
        \toprule
        \textbf{Method} & \textbf{Time} & \textbf{Gens} & \textbf{CFs (\%)} & \textbf{$Proximity$} & \textbf{$Sparsity$} \\
        \midrule
        \multicolumn{6}{c}{\texttt{German Credit}} \\
        p-value & $0.0\times 10^{0}$ & $3.5\times 10^{-20}$ & $3.9\times 10^{-1}$ & $5.4\times 10^{-72}$ & $3.4\times 10^{-3}$ \\
        Fix & $18.27$ (s) & $6.03$ & $95.36$ & $0.11$ & $0.02$ \\
        Random & $33.89$ (s) & $6.36$ & $96.07$ & $0.13$ & $0.02$ \\

        \midrule
        \multicolumn{6}{c}{\texttt{HELOC}} \\
        p-value & $0.0\times 10^{0}$ & $N/A$ & $7.5\times 10^{-1}$ & $0.0\times 10^{0}$ & $0.0\times 10^{0}$ \\
        Fix & $31.24$ (s) & $6.00$ & $92.43$ & $0.03$ & $0.03$ \\
        Random & $41.51$ (s) & $6.00$ & $92.16$ & $0.16$ & $0.02$ \\

        \midrule
        \multicolumn{6}{c}{\texttt{COMPAS}} \\
        p-value & $2.4\times 10^{-11}$ & $1.2\times 10^{-5}$ & $1.9\times 10^{-3}$ & $0.0\times 10^{0}$ & $2.2\times 10^{-44}$ \\
        Fix & $28.82$ (s) & $8.10$ & $47.86$ & $0.05$ & $0.03$ \\
        Random & $36.84$ (s) & $6.00$ & $44.31$ & $0.16$ & $0.04$ \\

        \midrule
        \multicolumn{6}{c}{\texttt{Adult}} \\
        p-value & $0.0\times 10^{0}$ & $7.5\times 10^{-7}$ & $9.7\times 10^{-14}$ & $0.0\times 10^{0}$ & $0.0\times 10^{0}$ \\
        Fix & $2.67$ (m) & $6.16$ & $79.55$ & $0.04$ & $0.02$ \\
        Random & $2.81$ (m) & $6.00$ & $75.49$ & $0.11$ & $0.03$ \\

        \midrule
        \multicolumn{6}{c}{\texttt{AdultCA}} \\
        p-value & $0.0\times 10^{0}$ & $1.2\times 10^{-75}$ & $5.2\times 10^{-1}$ & $0.0\times 10^{0}$ & $0.0\times 10^{0}$ \\
        Fix & $21.62$ (m) & $6.02$ & $57.35$ & $0.07$ & $0.06$ \\
        Random & $29.08$ (m) & $6.10$ & $57.48$ & $0.08$ & $0.05$ \\

        \bottomrule
    \end{tabular}
    \label{tab:fix_vs_random}
\end{table}

\subsection{Constraint Sequencing Analysis}
In this experiment, we evaluate whether the order in which user constraints are applied impacts the performance of UGCE-Incremental. Since users may impose constraints in arbitrary sequences during an interactive session, it is important to understand whether the ordering of updates affects counterfactual generation quality or efficiency.

To that end, we define three representative constraint sequences, each consisting of three update steps involving a different type of constraint:
\begin{enumerate}
    \item \textbf{I$\rightarrow$R$\rightarrow$D}: Immutability $\rightarrow$ Range $\rightarrow$ Directionality
    \item \textbf{R$\rightarrow$I$\rightarrow$D}: Range $\rightarrow$ Immutability $\rightarrow$ Directionality
    \item \textbf{D$\rightarrow$I$\rightarrow$R}: Directionality $\rightarrow$ Immutability $\rightarrow$ Range
\end{enumerate}
Each sequence is applied across all datasets using UGCE-Incremental. For every dataset and ordering, we report the total runtime for all instances, the counterfactual success rate (CFEs \%), and average quality metrics (\textbf{$proximity$} and normalized \textbf{$sparsity$}). While an exhaustive study of all constraint permutations is infeasible, this controlled setup provides a first look into whether the constraint sequence affects the behavior of UGCE-Incremental.

Table~\ref{tab:order_robustness} summarizes the results. Across all datasets and orderings, the differences in runtime and counterfactual quality are minimal. This suggests that UGCE is relatively insensitive to constraint sequencing, producing comparable outcomes regardless of the order in which constraints are introduced. An exception is observed in \texttt{AdultCA}, where \textbf{I$\rightarrow$R$\rightarrow$D} and then \textbf{R$\rightarrow$I$\rightarrow$D} sequences show better \textbf{$proximity$} and \textbf{$sparsity$} values and less total time.

Overall, the findings indicate that UGCE provides stable performance under different user interaction patterns, an important property for real-world deployment where constraint order may vary unpredictably.
\begin{table}[ht]
\centering
\caption{Constraint Ordering Comparison on UGCE-Incremental Method.}
\setlength{\tabcolsep}{3.5pt} 
\begin{tabular}{llccccc}
\toprule
\textbf{Dataset} & \textbf{Order} & \textbf{Time} & \textbf{CFs (\%)} & \textbf{$Proximity$} & \textbf{$Sparsity$} \\
\midrule
\multirow{3}{*}{\shortstack{\texttt{German}\\\texttt{Credit}}}
  & I→R→D & $1.29$ (m) & $98.81$ & $0.04$ & $0.02$ \\
  & R→I→D & $1.22$ (m) & $98.81$ & $0.04$ & $0.02$ \\
  & D→I→R & $1.36$ (m) & $98.81$ & $0.04$ & $0.02$ \\
\midrule
\multirow{3}{*}{\texttt{HELOC}}
  & I→R→D & $1.79$ (m) & $99.46$ & $0.05$ & $0.03$ \\
  & R→I→D & $1.72$ (m) & $99.46$ & $0.05$ & $0.03$ \\
  & D→I→R & $1.80$ (m) & $99.46$ & $0.05$ & $0.03$ \\
\midrule
\multirow{3}{*}{\texttt{COMPAS}}
  & I→R→D & $3.33$ (m) & $95.98$ & $0.04$ & $0.03$ \\
  & R→I→D & $3.15$ (m) & $93.41$ & $0.05$ & $0.03$ \\
  & D→I→R & $3.13$ (m) & $93.37$ & $0.05$ & $0.03$ \\
\midrule
\multirow{3}{*}{\texttt{Adult}}
  & I→R→D & $7.62$ (m) & $79.87$ & $0.03$ & $0.02$ \\
  & R→I→D & $7.28$ (m) & $80.04$ & $0.03$ & $0.02$ \\
  & D→I→R & $7.32$ (m) & $80.11$ & $0.03$ & $0.02$ \\
\midrule
\multirow{3}{*}{\texttt{AdultCA}}
  & I→R→D & $144.71$ (m) & $99.99$ & $0.06$ & $0.05$ \\
  & R→I→D & $170.42$ (m) & $99.99$ & $0.06$ & $0.05$ \\
  & D→I→R & $173.55$ (m) & $99.99$ & $0.10$ & $0.06$ \\
\bottomrule
\end{tabular}
\label{tab:order_robustness}
\end{table}

\subsection{Constraint Type Analysis}
To understand the impact of individual constraint types, we run UGCE-Incremental with only one constraint, \textit{immutability}, \textit{range}, or \textit{directionality}, enabled at a time. In all configurations, the constraint is applied to the \textit{same feature} across types to ensure a fair comparison. Table~\ref{tab:single_constraint_runs} reports the total runtime, average counterfactual success rate (CF \%), average \textbf{$proximity$}, and average normalized \textbf{$sparsity$} over five runs.

The results show no constraint type is universally easier or harder to satisfy. \textbf{Immutability constraints} yield the highest CF success on \texttt{German Credit}, \texttt{HELOC}, and \texttt{AdultCA}, but perform poorly on \texttt{COMPAS} and \texttt{Adult}. This suggests that the effect of immutability depends on the dataset and the role of the constrained feature.
\textbf{Range constraints} produce varied outcomes: strong on \texttt{German Credit} (64.88\%) but weak on \texttt{AdultCA} (39.51\%) and \texttt{Adult} (23.34\%), indicating their sensitivity to model boundaries and value distributions.
\textbf{Directionality constraints} consistently yield the lowest CF success, falling below 55\% in all datasets and as low as 15.94\% on \texttt{COMPAS}. Since these constraints enforce one-sided changes (e.g., only increases), they limit the search space and lead to both lower feasibility and slightly worse \textbf{$proximity$} and \textbf{$sparsity$}.
Runtime trends follow CF success: harder settings converge faster due to early failure, while feasible cases require more search effort. 

This experiment offers a preliminary lens into constraint difficulty. To draw broader conclusions, future work should evaluate constraint types across multiple features and combinations. These results underscore the need for dynamic, constraint-aware methods like UGCE to flexibly adapt to varying user feasibility demands.
\begin{table}[ht]
\centering
\caption{Single Constraint Type Comparison in UGCE-Incremental.}
\label{tab:single_constraint_runs}
\setlength{\tabcolsep}{2.4pt}
\begin{tabular}{lcccccc}
\toprule
\textbf{Dataset} & \textbf{Constraint Type} & \textbf{Time} & \textbf{CF (\%)} & \textbf{$Proximity$} & \textbf{$Sparsity$} \\
\midrule
\multirow{3}{*}{{\shortstack{\texttt{German}\\\texttt{Credit}}}} 
& Immutability   & 28.02 (s) & 92.86 & 0.06 & 0.02 \\
& Range          & 20.96 (s) & 64.88 & 0.08 & 0.02 \\
& Directionality & 16.45 (s) & 53.57 & 0.07 & 0.03 \\
\midrule
\multirow{3}{*}{\texttt{HELOC}}
& Immutability   & 36.50 (s) & 96.22 & 0.05 & 0.03 \\
& Range          & 20.16 (s) & 52.16 & 0.06 & 0.03 \\
& Directionality & 18.52 (s) & 49.46 & 0.06 & 0.03 \\
\midrule
\multirow{3}{*}{\texttt{COMPAS}} 
& Immutability   & 24.79 (s) & 33.70 & 0.04 & 0.03 \\
& Range          & 1.33 (m) & 61.05 & 0.06 & 0.03 \\
& Directionality & 12.56 (s) & 15.94 & 0.05 & 0.04 \\
\midrule
\multirow{3}{*}{\texttt{Adult}} 
& Immutability   & 53.56 (s) & 22.81 & 0.05 & 0.03 \\
& Range          & 59.93 (s) & 23.34 & 0.03 & 0.02 \\
& Directionality & 1.28 (m) & 36.85 & 0.04 & 0.02 \\
\midrule
\multirow{3}{*}{\texttt{AdultCA}} 
& Immutability   & 47.64 (m) & 97.95 & 0.14 & 0.06 \\
& Range          & 21.08 (m) & 46.24 & 0.10 & 0.06 \\
& Directionality & 17.84 (m) & 34.03 & 0.11 & 0.07 \\
\bottomrule
\end{tabular}
\end{table}
\section{Related Work}
\label{sec:relatedwork}
Explanations have become central in machine learning research 
\cite{fragkathoulas2024explaining,guidotti2024counterfactual}, particularly in high-stakes domains such as healthcare and education. Among various explanation methods, CFs have gained prominence for their ability to reveal actionable changes leading to a desired outcome.
Wachter et al. \cite{wachter2017counterfactual} first introduced CFEs as small, interpretable perturbations to input features that alter the decision of a model, offering insights into model behavior.
To refine CFEs, various methods \cite{guidotti2019factual,mothilal2020explaining,kanamori2020dace,goethals2024precof,sharma2020certifai,poyiadzi2020face}
have been proposed to enhance counterfactual generation, focusing on properties such as \emph{actionability} \cite{poyiadzi2020face}, \emph{sparsity} and divesity \cite{mothilal2020explaining}, \emph{robustness} \cite{guyomard2023generating}, and \emph{feasibility} \cite{poyiadzi2020face, mothilal2020explaining}.
Various methods optimize counterfactual search using \emph{genetic algorithms} (GAs) \cite{sharma2020certifai, guidotti2019factual, mothilal2020explaining}, \emph{integer programming} \cite{russell2019efficient,ustun2019actionable,kanamori2020dace}, and \emph{cost-based heuristics} \cite{goethals2024precof}, assuming a static feasibility space. 

However, in many real-world settings, the feasibility constraints of a user about a presented solution evolve over time, turning counterfactual generation into a dynamic optimization problem (DOP). 
As users iteratively update preferences (e.g., by disallowing changes to certain features or narrowing valid ranges), the feasible region shifts over time, requiring counterfactual methods that can adapt accordingly.
While a naive strategy would restart the search from scratch after each user update, this approach is inefficient and disregards valuable information gathered in earlier iterations \cite{branke2003designing}. Instead, several lines of work have focused on equipping Genetic Algorithms (GAs) with mechanisms to adapt efficiently in dynamic environments without full reinitialization.

Early approaches addressed adaptation in dynamic environments by enriching the genetic representation or altering population dynamics. Goldberg and Smith \cite{goldberg1987nonstationary} proposed encoding redundant genetic material, so-called \emph{diploid representations with dominance}, to retain alternative solutions that could become useful after environmental changes. Cobb \cite{cobb1990investigation} introduced \emph{hypermutation}, a mechanism that temporarily increases mutation rates to escape local optima when performance deteriorates. To maintain diversity, \emph{random immigrants} strategies periodically introduce new individuals into the population, as explored by Grefenstette \cite{grefenstette1992genetic} and Vavak et al. \cite{vavak1996comparative}. In contrast, \emph{memory-based methods} maintain a repository of high-performing solutions from past environments, which are reintroduced when similar conditions reoccur \cite{branke1999memory, louis1996genetic, mori2001adaptation}. Ramsey and Grefenstette \cite{ramsey1993case} 
says that environmental changes can be explicitly detected and classified.
Their architecture consists of an explicit environment monitoring, which restarts learning when change is detected, but leverages stored strategies to reinitialize the population more effectively. More recently, Yang \cite{yang2005memory} proposed a hybrid scheme 
that combines the advantages of memory and diversity. Immigrants are generated by mutating the best solution stored in memory and used to replace the worst individuals in the population.
This targeted diversity keeps the population biased toward the current environment and leads to more efficient adaptation than using either mechanism alone.

UGCE builds on these principles by applying population reuse in a simpler yet interactive setting. Unlike prior work that relies on environmental drift detection, UGCE operates in a user-guided context where the trigger for adaptation is explicit: the dissatisfaction of the user with a proposed counterfactual. When constraints are updated, UGCE avoids full reinitialization by warm-starting the search from the previously evolved population and selectively modifying only the individuals, or specific features, that violate the new constraints. This enables fast, constraint-aware adaptation across iterations without external monitoring mechanisms.
\section{Summary}
\label{sec:summary}
UGCE introduces a novel framework for incremental counterfactual explanation, tailored for real-world, interactive settings where users iteratively revise their feasibility constraints. Unlike prior work, most of which regenerates explanations from scratch after every constraint change, UGCE retains and repairs the previous solutions, enabling efficient, dynamic optimization.
The method builds on Genetic Algorithms (GAs) and extends them with a warm-start mechanism that repairs constraint-violating individuals instead of restarting the entire population. A key innovation is the ability of UGCE to reuse prior knowledge, adapting seamlessly to new user constraints like immutability, range bounds, and directionality.
Through comprehensive experiments across five benchmark datasets, UGCE is shown to:
(a) Drastically reduce computation time (especially in large-scale or iterative cases),
(b) Maintain high-quality, not sparse, and proximal counterfactuals,
(c) Demonstrate stability across runs,
(d) Perform well under varying sequences of user constraint updates.

\section{Future Directions}
\label{sec:future_directions}
Future directions include enhancing UGCE with dynamic weighting schemes that adjust trade-offs (e.g., between sparsity and proximity) based on user interaction history and evolving preferences. Another promising direction is incorporating feedback signals, such as user approvals, corrections, or relevance ratings, into the optimization loop to personalize counterfactuals. UGCE could also benefit from domain-specific priors or feasibility rules derived from expert knowledge or empirical distributions. Finally, deploying UGCE in real-time interactive settings with longitudinal user engagement would assess its practical impact.


\section{Acknowledgment}
This work has been partially supported by project MIS 5154714 of the National Recovery and Resilience Plan Greece 2.0 funded by the European Union under the NextGenerationEU Program.

\bibliographystyle{IEEEtran}
\bibliography{bibliography_mid}

\end{document}